\documentclass{article}
\usepackage{ijcai21}

\usepackage{times}

\usepackage{soul}
\usepackage{url}
\usepackage[dvipsnames]{xcolor}
\usepackage[hidelinks]{hyperref}
\usepackage[utf8]{inputenc}
\usepackage{xspace}
\usepackage{multirow}
\usepackage{booktabs}
\usepackage[small]{caption}
\usepackage{graphicx}
\usepackage{amsmath}
\usepackage{amssymb}
\usepackage{booktabs}
\usepackage{todonotes}
\usepackage{bbm}
\usepackage{amsthm}
\newtheorem{definition}{Definition}[section]

\usepackage{color} % Color

\title{Learning Neighborhood Representation from Multi-Modal Multi-Graph: Image, Text, Mobility Graph and Beyond}
\author{
Tianyuan Huang\footnote{Equal contribution}$^1$\and
Zhecheng Wang\footnotemark[1]$^1$\and
Hao Sheng\footnotemark[1]$^1$\and
Andrew Y. Ng$^1$ \And
Ram Rajagopal$^1$\\
\affiliations
$^1$Stanford University
\emails
\{tianyuah, zhecheng, haosheng, ramr\}@stanford.edu, ang@cs.stanford.edu
}

\begin{document}

\maketitle

\newcommand{\method}{M3G\xspace}
\newcommand{\zhecheng}[1]{\textcolor{red}{\textbf{[Zhecheng: #1]}}}
\newcommand{\hao}[1]{\textcolor{blue}{\textbf{[Hao: #1]}}}
\newcommand{\tianyuan}[1]{\textcolor{orange}{\textbf{[Tianyuan: #1]}}}
\newcommand{\td}[1]{\textcolor{red}{\textbf{#1}}}

\begin{abstract}
Recent urbanization has coincided with the enrichment of geotagged data, such as street view and point-of-interest (POI). Region embedding enhanced by the richer data modalities has enabled researchers and city administrators to understand the built environment, socioeconomics, and the dynamics of cities better. 
While some efforts have been made to simultaneously use multi-modal inputs, existing methods can be improved by incorporating different measures of ``proximity'' in the same embedding space --- leveraging not only the data that characterizes the regions (e.g., street view, local businesses pattern) but also those that depict the relationship between regions (e.g., trips, road network).
To this end, we propose a novel approach to integrate multi-modal geotagged inputs as either node or edge features of a multi-graph based on their relations with the neighborhood region (e.g., tiles, census block, ZIP code region, etc.). We then learn the neighborhood representation based on a contrastive-sampling scheme from the multi-graph. 
Specifically, we use street view images and POI features to characterize neighborhoods (nodes) and use human mobility to characterize the relationship between neighborhoods (directed edges).
We show the effectiveness of the proposed methods with quantitative downstream tasks as well as qualitative analysis of the embedding space: The embedding we trained outperforms the ones using only unimodal data as regional inputs.
\end{abstract}

\section{Introduction}
% Paragraph 1: the world is full of multi-modal graphs such as websites and urban neighborhoods. Each object in the graph is a ``container'' containing multiple sub-objects like images and text, and there are various connections between objects. How to represent the objects in multi-modal graphs?
\begin{figure}[ht!]
  \centering
  \makebox[0pt]{
  \hspace{-45pt}
  \includegraphics[width=0.55\textwidth]{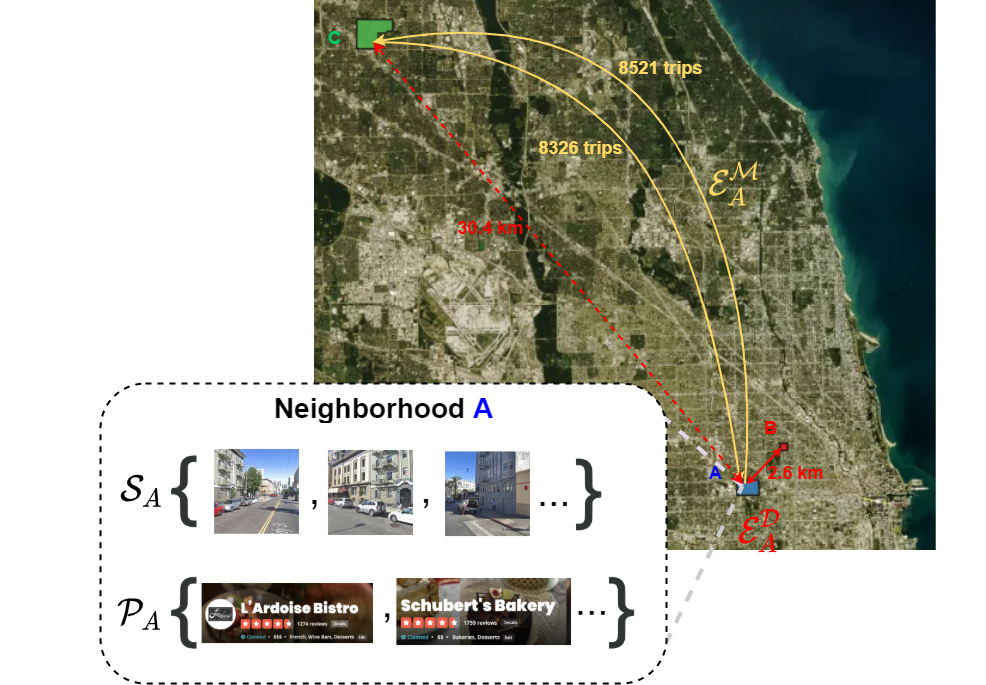}}
  \caption{Multi-modal multi-graph of urban neighborhoods in the City of Chicago. Each neighborhood is a {\it container} of multi-modal inputs, e.g., street views and POIs. The neighborhoods are considered connected if they are close spatially (e.g. {\color{BlueViolet} A} and {\color{Red}B}) or or if there are many human mobility trajectories in between (e.g. {\color{BlueViolet} A} and {\color{ForestGreen}C}). 
  %Here the mobility measures are irreciprocal and we model them as directed edges. 
  Notice that even {\color{BlueViolet} A} and {\color{ForestGreen}C} are spatially far away, the large number of trips in between indicates the strong relations which should be captured in the embedding space.}
  \label{fig:neighborhoods}
\end{figure}
The world is full of connections between entities of different modalities, such as websites and urban neighborhoods. 
A website can be represented as a node containing multi-modal components like text, images, and videos;  hyperlinks connected websites as directed edges. 
Similarly, an urban neighborhood can be regarded as a complex multi-modal node containing the natural and built environment, business activities, and the people living there. 
Urban neighborhoods are interconnected implicitly with various types of relations such as geospatial proximity and human mobility trajectories between neighborhoods.
With the vision of ``smart city'' being proposed in different parts of the world as well as the increasing availability of a great variety of data in cities, understanding the characteristics and dynamics of cities become essential, and more importantly, feasible with the help of state-of-the-art machine learning algorithms.
Urban neighborhood embedding, or representing various urban features as vectors, is a preliminary task to many data-driven urban studies and applications such as spatiotemporal prediction, planning, and causal inference.
Though abundant studies focus on representation learning for a single modality of data like images~\cite{radford2015unsupervised} and text~\cite{mikolov2013efficient}, representing urban neighborhoods leveraging multi-modal data while maintaining their correlations is still a challenging task.

% Paragraph 2: (1) Represent  with census data? Expensive data collection, and limited by the pre-defined and inflexible aggregated level (e.g. zipcode, county, etc) (2) Represent it with supervised learning? Inflexible, limited transferability, label-hungry. Conclusion: unsupervised representation learning are required.
Traditional data collection approaches like census are costly.
For example, the 2020 U.S. census was estimated to cost 15.6 billion dollars ~\cite{Census}. 
Moreover, the data produced by census is usually aggregated at pre-defined geographic divisions (e.g., census tracts and counties) and can hardly be re-mapped into other customized geospatial units such as raster tiles or polygons, which limits the flexibility of using the data.
There are recent attempts to extract or predict urban characteristics from widely-available urban-associated data using data-driven approaches, including both supervised and unsupervised learning. 
Supervised learning methods utilize geo-tagged data such as point-of-interest (POI) \cite{yuan2012discovering}, and street view imagery \cite{gebru2017using} as inputs and output the inference of local socioeconomic attributes.
However, supervised learning is task-specific: The representation learned is not necessarily transferrable to other tasks.
% Though the representation of urban neighborhoods can be learned implicitly, it is not necessarily transferrable to other tasks.
Furthermore, developing supervised learning models with high-dimensional data like images requires a massive dataset with annotated labels of ground-truth socioeconomic attributes, which is not necessarily available for certain regions or at the desired geographic level (e.g., raster tiles).
By contrast, unsupervised learning overcomes such limitations by developing a universal and versatile representation without task-specific ground-truth supervisions.
Common urban features to use include POI  \cite{fu2019efficient}, street views \cite{law2019unsupervised}, and taxi trips \cite{yao2018representing}.
However, most of the existing unsupervised urban representation learning is still based on unimodal data, without fully leveraging various types of data both within and between neighborhoods.

% Paragraph 3: The nature of urban neighborhoods is multi-modal graphs consisting of image, text, and connections such as mobility or geospatial proximity. Need the integration of multiple modalities.

% Paragraph 4: To learn the representation of multi-modal graph, we proposal the urban2vec+, a three-stage unsupervised learning framework to incorporate street view imagery, POI text data, and mobility connection.

Urban neighborhoods are complex systems that can be modeled by a multi-modal multi-graph: Each urban neighborhood (``node'')
is a ``container'' which contains the built environment,  business activities, and population inside the neighborhood. There are also relations (``edge'') between neighborhoods, which can be characterized by geospatial proximity, mobility connections, or both.
To obtain a comprehensive representation of urban neighborhoods, we model the neighborhoods in an urban area as multi-modal multi-graph (\method) and develop an unsupervised representation learning framework to obtain the neighborhood embedding from the graph.
% Specifically, we leverage % the widely-available
% street view images and POI features to capture the intra-neighborhoods characteristics, and use geospatial distance and mobility patterns between neighborhoods to learn their mutual relations.
%To extract semantics and correlations from different modalities of data, we develop a three-stage learning framework, with each stage integrating one modality of data into the neighborhood embedding:
%For street view images, we use convolutional neural network to extract visual features; 
%for POI, we dismantle the POI information such as price, rating, and reviews into bag of words and correlates the POI word embedding with the neighborhood embedding; 
% We construct triplet loss to provide constrastive samples 
Instead of learning the graph globally, we propose a contrastive sampling approach that samples triplet (anchor, positive, negative) according to the multi-graph edges, enabling scalable training with multi-city data.  
%The first two stages integrate street views and POIs into the representation of the neighborhood as a ``container'' while the last stage further establish the neighborhood-neighborhood relations in the embedding space. 
Our major contribution is three-fold: 1) We proposed a framework to learn neighborhood representation by jointly modeling both inter- and intra-neighborhood multi-modal data as a multi-graph.
%Our major contribution is three-fold: 1) Our proposed framework learn the neighborhood representation by modeling the urban neighborhoods as a multi-modal multi-graph: In our implementation, ``nodes'' are ``containters'' containing the natural and built environment (street views) and business patterns (POIs), and ``edges'' are modeled by geospatial proximity and/or mobility flows. Such framework can be easily to incorporate other types of data such as satellite imagery and social media data.
2) We demonstrate this framework with real-world data in two U.S. metropolitan areas at the census-tract level, using street view images and POI features as intra-neighborhood characteristics, and geospatial proximity and mobility flow as inter-neighborhood relations. The neighborhood embeddings generated from our framework achieve state-of-the-art performance in all downstream prediction tasks.
%2) Our experiments on three U.S. metropolitan areas demonstrate the state-of-the-art performance of the generated neighborhood embedding in downstream prediction tasks.
%3) We qualitatively show the proximity between neighborhood (``container'') embedding and the embeddings of street views it contains, as well as the proximity between two neighborhood embeddings in the embedding space that varies with their mobility connections.
%Such analysis shows that our model successfully integrates different modalities of data in the embedding space.
3) We propose three qualitative evaluations for the neighborhood embedding space, showing that our model successfully integrates various data modalities in the embedding space.

% Our contribution is three-fold: (1) a three-stage learning framework to capture three different modalities within and between neighborhoods as well as their relationships.\hao{Could high-light more the modality on edges as well as on nodes, and mention we treat the negihborhood as a container of multi-modalities} (2) achieved state-of-the-art performance in downstream prediction tasks. (3) Present a novel way (conjugated key+query embeddings) to learn the neighborhood representation from mobility data.\hao{Would suggest excange 2 and 3} 

\section{Related Work} 
\subsubsection{Spatiotemporal Representation Learning}
% Transfer learning in NLP and CV have shown the power of pre-trained model ~\cite{mahajan2018exploring,brown2020language}.
Spatiotemporal representation learning aims to produce region embedding using geo/temporal-tagged data under the First Law of Geography~\cite{tobler1970computer}\footnote{``Everything is related to everything else, but near things are more related than distant things.''}. 
\cite{chu2019geo,mac2019presence} generate geo-aware prior based on the geo-coding of coordinates. 
Tile2vec~\cite{jean2019tile2vec} starts the stream of imposing such prior to the embedding space through contrastive learning. 
Using geo-proximity as the single criterion to sample positive and negative tiles, this algorithm judiciously pushes the latter further away from the anchor point in the embedding space as compared with the former.
Unfortunately, such framework can not be easily applied to multi-modal settings as a consistent and meaningful distance measure is required between any two samples across different modalities. 
Urban2Vec~\cite{wang2020urban2vec} overcomes such drawbacks by introducing the neighborhood embedding. 
It is worth noticing the spatiotemporal relation between each sample can be viewed as a reciprocal relation denoted by an undirected edge. 
\cite{jiang2020survey} introduces the use of mobility, POI similarity or even the likeness of geo-tagged tweets~\cite{zhang2017regions} as new metrics of proximity to define ``edges''.  
In this work, we generalize the contrastive learning approach to non-reciprocal relations such as mobility flow and propose a framework that can be easily extended to other graph-structured datasets with multi-modal edges and multi-modal nodes. 

\subsubsection{Graph Embedding}
There are a lot of graph embedding methods (e.g., DeepWalk~\cite{perozzi2014deepwalk}, node2vec~\cite{grover2016node2vec}) that generates embedding for a certain node in the graph. 
They can be applied to the mobility graph.
For example, \cite{fu2019efficient} incorporate such prior by directly impose an autocorrelation in the latent space. 
However, most of them are not able to model multi-modal edge (as in a multi-graph), and their embedding space does not reflect the multi-perspective proximity between nodes. 
To further incorporate information from both nodes (e.g. POI, street view) and edges (e.g. mobility, distance), \cite{jenkins2019unsupervised} concatenate image embedding and graph embedding at each node. 
Our training strategy can be viewed as an extension of the contrastive sampling technique in Graph Neural Network setting (\cite{schroff2015facenet,qiu2020gcc}): By sampling triplets according to multiple proximity measures, the embedding captures the multi-graph topological properties as well as the multi-modal features from each node. 
 
\subsubsection{Urban Computing}
Urban Computing aims to tackle major issues in cities, such as traffic control, public health and economic development, by modeling and analyzing urban data. 
A lot of research have shown the possibility to infer this socioeconomic information from satellite image~\cite{jean2016combining,Sheng_2020_CVPR_Workshops}, street view~\cite{gebru2017using}, human mobility~\cite{xu2018human} and geo-tagged social network activities~\cite{schwartz2014social}. 
Recent studies also demonstrate that similar tasks could benefit from %more than 
multi-modal inputs: ~\cite{wang2016crime} utilizes both POI
data and taxi trip data to infer crime rate in Chicago. 
%~\cite{wegner2016cataloging} matched the detections from both satellite image and street view to better catalog and localize the trees in Pasadena, CA, USA.
~\cite{irvinforestnet} includes a fusion of auxiliary variables, such as elevation and air pressure, with a computer vision model on satellite images to improve the performance of forest loss driver classification. 
We hope the multi-graph framework proposed in this work will provide a much convenient and comprehensive tool for urban computing tasks with multi-modal data.  

%City environment~\cite{mooney2016use}; Street score~\cite{naik2014streetscore}.
%Housing price~\cite{law2019take}.
%Covid19~\cite{nguyen2020using,kang2020multiscale}.

\section{Methods}
In the following section, we first mathematically define the problem of learning neighborhood embedding and give an overview of the construction of Multi-Modal Multi Graph (\method). Then we introduce the concept of {\it neighborhood container} and our contrastive sampling strategy to incorporate multi-modal inputs at each node. We continue by describing our inter-neighborhood learning strategy for both directed and undirected edges. This section is concluded by a summary of the loss function used in \method. 

\subsection{Problem Statement}
Unlike most of the previous studies that 
% are solving the problem for 
focus on specific modality (e.g., image, text, etc.) and specific geographic unit (e.g. census tract, county, etc.), we restate the general problem of Urban Neighborhood Embedding agnostic to both as the following:  
\begin{definition}[Urban Neighborhood Embedding Problem]
Given a metropolitan area $\mathcal{A}$ that is composed of a set of disjointed neighborhood geometries $\mathcal{U} = \{u_1, u_2, ..., u_N\}$, s.t. $\mathcal{A} = \bigcup_i^N u_i$, the goal of urban neighborhood embedding is to learn a vector representation $z_i \in \mathbb{R}^d$ for each $u_i$ which encodes the characteristics and mutual relations of $u_i$.
\end{definition}
Notice $u_i$ can be a raster tile of certain size (commonly used in remote sensing), a census tract or a county. Under our abstraction we do not assume all $u_i$ are of the same geographic unit. \\

Geo-tagged data (i.e. data with GPS coordinates) is used to generate such embedding. Instead of categorising data by the modality, we use a more general approach of categorization based on how data is associated with the location(s):

\begin{definition}[Geo-Tagged Point Data]
Geo-tagged point data is the kind of data characterizing one geolocation $l$:
\[
\mathcal{D}^{p}_m = \{(x^m, l)\}
\] is the set of geo-tagged point data with an input $x^m$ of modality $m$ at each geolocation. Examples of geo-tagged point data includes street views, POI check-in data and satellite images.
\end{definition}

\begin{definition}[Geo-Tagged Reciprocal Data]
Geo-tagged reciprocal data is the kind of data characterizing the relation between two geolocations $l_1$ and $l_2$, but it does not have a direction and the relation is reciprocal:
\[
\mathcal{D}^{r}_m = \{(x^m, l_1, l_2)\}\bigcup\{(x^m, l_2, l_1)\}
\]
is the set of geo-tagged reciprocal data with an input $x^m$ of modality $m$ between two geolocations.
Examples of geo-tagged reciprocal data include spatial distance, road connectivity, and transaction volume.
\end{definition}

\begin{definition}[Geo-tagged Irreciprocal Data] \label{def: rec}
Geo-tagged reciprocal data is the kind of data characterizing the relation between two geolocations $l_1$ and $l_2$ with a direction: 
\[
\mathcal{D}^{ir}_m = \{(x^m, l_1, l_2)\}
\]
is the set of geo-tagged irreciprocal data with an input $x^m$ of modality $m$ between two geolocations.
Examples of geo-tagged irreciprocal data include human mobility, commute time, and goods imports/exports.
\end{definition}
The three categories of data are corresponding to the node, undirected, and directed edges in our \method model and will be further explained in the next two sections. 
For now, let us assume $\mathcal{D} = \bigcup_{m, t}\mathcal{D}^{t}_m$ and introduce the concept of multi-modal multi-graph:

\begin{definition}[Multi-Modal Multi-graph (\method)]
The Multi-Modal Multi-graph $\mathcal{G}_{\mathcal{D}}(\mathcal{U},\mathcal{E})$ is a multi-graph for neighborhoods $\mathcal{U}$ and their edge set $\mathcal{E}$, characterized by the multi-modal geo-tagged dataset $\mathcal{D}$.
The nodes $\mathcal{U}$ have attributes defined by all geo-tagged points data $\mathcal{D}^{p}_m$, which are described with more details in Section \ref{sec: intra}. The edges $\mathcal{E}$ are defined by all geo-tagged reciprocal/irreciprocal data $\mathcal{D}^{r}_m$ and $\mathcal{D}^{ir}_m$, which are described in Section \ref{sec: inter}.
\end{definition}

\subsection{Intra-Neighborhood Modalities} \label{sec: intra}
Despite their vast difference in data structure, both POI meta information and street view images depict the urban characteristics at specific location. In this section, we will use them as examples of \textit{Intra-Neighborhood Modalities} and demonstrate how we incorporate their information into the neighborhood embedding. 

\subsubsection{Neighborhoods as Containers}
Given a set of geo-tagged street view images $\mathcal{D}_\mathcal{S}^{p} = \{(x^\mathcal{S}, l)\}$, where $s$ is an image and $l$ is its geolocation, we can easily assign each data point to the urban neighborhood $u_i$ it is located in: 
\[
\mathcal{S}_i = \{x^\mathcal{S} | (x^\mathcal{S}, l) \in \mathcal{D}_\mathcal{S}^{p} \text{, s.t. } l\in u_i\}
\]
Each $\mathcal{S}_i$ is a bag of street view images for neighborhood $u_i$.

Similarly, we can construct the feature container with the POIs $\mathcal{D}_\mathcal{P}^{p} = \{(x^\mathcal{P}, l)\}$, where $p$ is a POI and $l$ is its geolocation. To represent each POI $p$, we further disassemble the textual information of $p$, which are extracted from the POI category, price, and customer reviews, into a bag of words $\{t\}$. By pooling bags of words of all POIs inside a neighborhood, we obtain the bag of POI words for each neighborhood  $u_i$ in \method.
% Instead of merging all the text information directly,
\[
\mathcal{P}_i = \{t | (x^\mathcal{P}, l) \in \mathcal{D}_\mathcal{P}^{p} \text{, s.t. } t\in x^\mathcal{P} \text{ and } l\in u_i\}
\]
$t$ denotes a word. We can extend this approach to incorporate other textual data such as geo-tagged social media posts.

\subsubsection{Intra-Neighborhood Contrastive Learning Objective}
With the node feature containers $\mathcal{S}_i$ and $\mathcal{P}_i$ constructed, we here propose our intra-neighborhood contrastive-sampling strategy: For each pass, we sample one neighborhood $u_a$ uniformly at random from $\mathcal{U}$, i.e. $u_a \overset{\text{u}}{\sim} \mathcal{U}$, as our anchor neighborhood. Then we sample one context street view image $s_c\overset{\text{u}}{\sim}\mathcal{S}_a$ and one negative street view image $s_n \overset{\text{u}}{\sim}\mathcal{S}_{-a}$, with $\mathcal{S}_{-a}=\bigcup_{i\neq a} \mathcal{S}_{a}$.  Our proposed triplet loss~\cite{schroff2015facenet} formulates as:
\begin{align} \label{eq: loss_s}
    \mathcal{L}_{\mathcal{S}}(z_a, s_c, s_n) 
    = [M + ||z_a - f_\theta(s_c)||_2  - ||z_a - f_\theta(s_n)||_2]_+ 
\end{align}
, where $[\cdot]_+$ is a rectifier and a positive constant $M$ is used to prevent infinitely large difference between these two distances.
$z_a$ is the embedding vector for neighborhood $u_a$.
$f_\theta(\cdot)$ is the learnable encoder for images, e.g. a convolutional neural network with parameters $\theta$.

Similarly, given a random sample $u_a$ from $\mathcal{U}$, we can sample POI word $t_c\overset{\text{u}}{\sim}\mathcal{P}_a$ and $t_n\overset{\text{u}}{\sim}\mathcal{P}_{-a}=\bigcup_{i\neq a} \mathcal{P}_{a}$ and construct the triplet loss for POI data:
\begin{align} \label{eq: loss_p}
        \mathcal{L}_{\mathcal{P}}(z_a, t_c, t_n) = [M + ||z_a - g_\phi(t_c)||_2  - ||z_a - g_\phi(t_n)||_2]_+
    \end{align}
The definitions of $[\cdot]_+$ and $M$ are the same as above. $g_\phi(\cdot)$ is the learnable encoder for word with parameters $\phi$.
\begin{figure*}[ht!]
  \centering
  \includegraphics[width=1\textwidth]{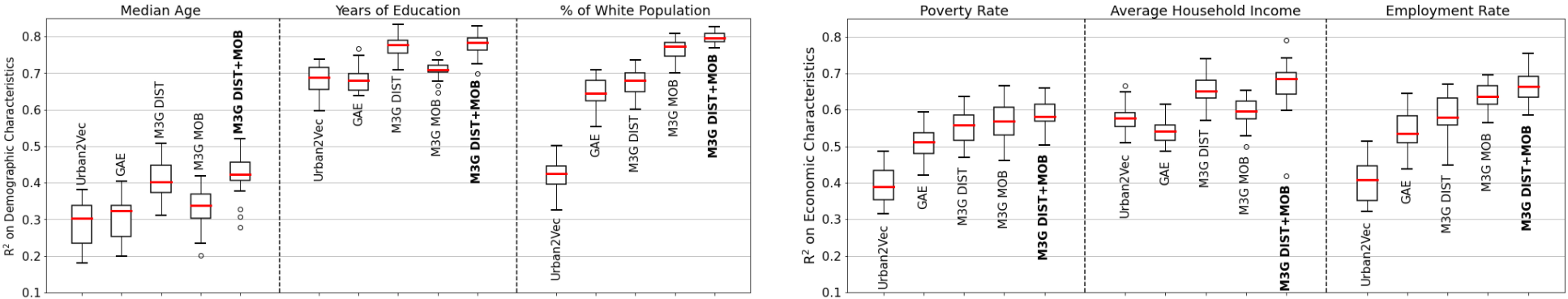}  
  \caption{Prediction $R^2$ on neighborhood attributes with random forest model in Chicago. {\bf Left}: Demographic attributes. {\bf Right}: Economic attributes.}
  \label{fig:plot}
\end{figure*}
\subsection{Inter-Neighborhood Modalities} \label{sec: inter}
Without data characterizing the relations between neighborhoods, the neighborhood embedding obtained by minimizing ~(\ref{eq: loss_s}) and (\ref{eq: loss_p}) can only incorporate information within neighborhoods ~\cite{wang2020urban2vec}. In this section, we will describe how $\mathcal{D}^{r}_j$ and $\mathcal{D}^{ir}_j$ characterizes the edges in graph $\mathcal{G}$ and introduce our learning strategy for inter-neighborhood modalities. We include both spatial distance $\mathcal{D}^{r}_\mathcal{D}$ and human mobility $\mathcal{D}^{ir}_\mathcal{M}$ as examples of inter-neighborhood modalities.

\subsubsection{Multi-Modal Multi-Edges}
Spatial distance can be measured between any pair of neighborhoods $(u_i, u_j)$. We can define the \textit{outgoing} edge sets of $u_i$ induced from the spatial distance as: 

\begin{align*}
    \mathcal{E}_i^\mathcal{D} = \{(u_i, &u_j, x^\mathcal{D}) | (x^\mathcal{D}, l_1, l_2) \in \mathcal{D}^{r}_\mathcal{D} \\ 
    &\text{ s.t. } l_1 \in u_i \text{ and } l_2 \in u_j\}
\end{align*}

Here $x^\mathcal{D} = \frac{1}{d_{ij}}$, which is the reciprocal of geospatial distance between $u_i$ and $u_j$.
Notice that $\mathcal{D}^{r}_\mathcal{D}$ already includes both directions of a same undirected edge according to Definition~\ref{def: rec}.
%we denote a weighted directed edge as $(start, end, value)$. 
Similarly we can define the \textit{outgoing} edge sets of $u_i$ induced from the human mobility $\mathcal{D}^{ir}_\mathcal{M}$:

\begin{align*}
    \mathcal{E}_i^\mathcal{M} = \{(u_i, &u_j, x^\mathcal{M}) | (x^\mathcal{M}, l_1, l_2) \in \mathcal{D}^{ir}_\mathcal{M} \\ 
    &\text{ s.t. } l_1 \in u_i \text{ and } l_2 \in u_j\}
\end{align*}

Here $x^\mathcal{M}$ is the total number of trips from a geolocation in $u_i$ to a geolocation in $u_j$.
Once we add both sets of edges to the graph $\mathcal{G}$, 
it is likely there can be multiple edges between $u_i$ and $u_j$ from different modalities.
%it is likely there can be multiple edges between $u_i$ and $u_j$: They could either come from different modalities or single modality but different samples (e.g., two records of taxi trips between $u_i$ and $u_j)$.

%Here we use mobility patterns data from SafeGraph, a company that aggregates anonymized location data from mobile devices. For each Point-of-interest (POI), the dataset contains the address information for this POI, as well as the counts of the number of visitors, and aggregated estimates of the home Census-block Groups (CBGs) of visitors. We locate the neighborhood $n_{i}$ by each POI's address, and the neighborhood $n_{j}$ from the counts of the number of visitors column. We assume there're $m$ trip between $n_{i}$ and $n_{j}$ if $m$ visitors' homes locate in neighbodhood $n_{j}$.
\subsubsection{Inter-Neighborhood Contrastive Learning Objectives}
Like Section \ref{sec: intra}, we first sample one neighborhood $u_a$ at random from $\mathcal{U}$, i.e. $u_a \overset{\text{u}}{\sim} \mathcal{U}$.
Instead of defining the context and negative set explicitly as in Section \ref{sec: intra},
%for intra-neighborhood modality, 
we draw samples of context neighborhood by sampling each edge with the probability proportional to the weights associated with it. 
Specifically, edge $(u, v, w)$ has weight of $p_m(w)$ being sampled, with $p_m(\cdot)$ a designed thresholding function using the prior on modality $m$. For example, for the spatial distance, we can set 
\[
    p_\mathcal{S}(w) = 
    \begin{cases}
    1, \quad \text{if } w > \frac{1}{500} \\
    0, \quad \text{otherwise}
    \end{cases}
\]
to sample a context neighborhood within a radius of 500 meters. 
Hence, for modality $m \in \{\mathcal{D}, \mathcal{M}\}$, the probability of $u_j$ being sampled as a context neighborhood $u_c$ is:
\begin{equation}
    P^m_{a,j} = \frac{\sum_{(u,v,w) \in \mathcal{E}_a^m } p_m(w) \mathbbm{1}_a(u)\mathbbm{1}_j(v)}{\sum_{(u, v, w) \in \mathcal{E}_a^m } p_m(v)\mathbbm{1}_a(u)}
\end{equation}
Here $\mathbbm{1}_x(\cdot)$ is the indicator function with the value 0 everywhere except for $x$.
The negative neighborhood $u_n$ is sampled uniformly at random from the set of rest of nodes $\{u_j|P_{a,j}^m = 0\}$. 
Finally, we have the inter-neighborhood triplet loss for each modality $m \in \{\mathcal{D}, \mathcal{M}\}$: 
\begin{align} \label{eq: loss_n}
    \mathcal{L}_{m}(z_a, z_c, z_n) = [M + ||z_a - z_c||_2  - ||z_a - z_n||_2]_+
\end{align}
The definitions of $[\cdot]_+$ and $M$ are the same as above. By default, we sample balanced number of triplets for each modality. Together with Equation~(\ref{eq: loss_s}) and (\ref{eq: loss_p}), we are able to train our neighborhood embedding with any modality of inter/intra-neighborhood data. Next section will demonstrate our framework with experiments on real-world datasets.

\section{Experiment}
\begin{table*}[t!]
    \centering
    \resizebox{1\textwidth}{!}{%
    \begin{tabular}{l|c|c|c|c|c|c}
      \toprule
        \multirow{2}{*}{Model} & \multicolumn{3}{|c}{Demographic characteristics} & \multicolumn{3}{|c}{Economic characteristics} \\
          \cmidrule{2-7} 
        &  Median age & Years of education & Percentage of white population & Poverty rate & Average household income & Employment rate\\ 
      \midrule
      Urban2Vec \cite{wang2020urban2vec}& $0.326\pm0.056$ & $0.701\pm0.035$  & $0.472\pm0.052$ & $0.418\pm0.052$ & $0.515\pm0.052$ & $0.441\pm0.059$\\
      GAE \cite{DBLP:journals/corr/KipfW16}& $0.261\pm0.072$ & $0.672\pm0.024$ & $0.480\pm0.061$ & $0.432\pm0.078$& $0.457\pm0.047$ & $0.435\pm0.079$\\
      \method DIST & $0.344\pm0.052$ & $0.756\pm0.030$ & $0.630\pm0.038$ & $0.488\pm0.053$ & $0.548\pm0.047$ & $0.530\pm0.046$\\
      \method MOB & $0.338\pm0.063$ & $0.780\pm0.020$ & $\mathbf{0.736\pm0.021}$& $0.591\pm0.049$ & $0.616\pm0.029$ & $0.615\pm0.038$\\
      \method DIST+MOB & $\mathbf{0.374\pm0.060}$ & $\mathbf{0.790\pm0.022}$ & $0.734\pm0.030$ & $\mathbf{0.602\pm0.049}$ & $\mathbf{0.630\pm0.038}$ & $\mathbf{0.627\pm0.036}$\\
      \bottomrule
    \end{tabular}
    }
    \caption{Prediction $R^2$ on demographic and economic attributes with linear regression model in Chicago.}
    \label{table:demographics}
\end{table*}
To demonstrate the effectiveness of our framework, we conduct experiments on 1294 census tracts in Chicago and 1310 census tracts in New York City. We demonstrate our framework at census-tract level because the reference data for prediction (e.g., American Community Survey (ACS)) are readily available at this level. Our framework can be easily applied to other geographic divisions (e.g. block groups) or even customized units (e.g. raster tiles).

\subsection{Data Description}
% Here we use mobility patterns data from SafeGraph, a company that aggregates anonymized location data from mobile devices. For each Point-of-interest (POI), the dataset contains the address information for this POI, as well as the counts of the number of visitors, and aggregated estimates of the home Census-block Groups (CBGs) of visitors.
The street view images and POI features we used are obtained from Google Street view API\footnote{\url{https://developers.google.com/maps/documentation/streetview}} and Yelp Fusion API\footnote{Available at \url{https://www.yelp.com/fusion}}, respectively. 
We randomly sample 50 street views for each census tract. 
The human mobility data is provided by SafeGraph\footnote{See data catalog  at \url{https://docs.safegraph.com/docs/}.}. Specifically, we use Core Places and Weekly Patterns datasets, which include, for each POI, the exact location, as well as the aggregated weekly estimates of the home CBGs of visitors. We preprocess the weekly patterns in Chicago and New York City from Jan 2018 to Dec 2019. Each visit is encoded as a directed edge between neighborhoods of POI and visitor's home; both are aggregated at the census tract level.
Their statistics are summarized in Table~\ref{table:graph}.

\begin{table}[h!]
    \centering
    \resizebox{0.45\textwidth}{!}{%
    \begin{tabular}{l|c|c|c}
      \toprule
       & \ Area ($km^2$)  & \# Edges & Average in/out degree\\ 
      \midrule
      Chicago & $606$ & $143,235$ & $110$ \\ 
      New York City & $1212$ & $120,470$ & $92$ \\
      \bottomrule
    \end{tabular}
    }
    \caption{Safegraph mobility data statistics}
    \label{table:graph}
\end{table}

% \begin{table}[h!]
%     \centering
%     \resizebox{0.5\textwidth}{!}{%
%     \begin{tabular}{l|c|c|c|c}
%       \toprule
%       & \# Street views & \# POIs & \# Trips & \# Neighborhoods\\ 
%       \midrule
%       Chicago & $64,739$ & $38,445$ & $732,697$ & $1,294$\\ 
%       New York City & $67,271$ & $50,697$ & $606,696$ & $1,310$\\
%       \bottomrule
%     \end{tabular}
%     }
%     \caption{Dataset statistics}
%     \label{table:setup}
% \end{table}
\subsection{Training Details}
For all experiments we set embedding dimension $d=200$ for images, POI words, and neighborhood. 
We use an Inception-v3~\cite{Szegedy_2016_CVPR} architecture as the encoder for street view images (i.e., $f_\theta(\cdot)$ in Equation~(\ref{eq: loss_s})). 
%It has final linear layer projecting into a 200-dimensional embedding space. 
The encoder for POI words(i.e., $g_\phi(\cdot)$ in Equation~(\ref{eq: loss_p})) is a look-up table with weights initialized by GloVe~\cite{pennington-etal-2014-glove}.
During training, we minimize loss ~(\ref{eq: loss_s}), ~(\ref{eq: loss_p}), ~(\ref{eq: loss_n}) sequentially in a three-stage process. When we sample inter-neighborhood triplet, for spatial distance, we sample $u_c$ uniformly at random from the $5$ closest neighbors and sample $u_n$ uniformly at random from the rest. %For human mobility,  $u_c$ is sampled from all adjacent node with a probability proportional to the weights of the edge in between and $u_n$ is sampled uniformly at random from the rest. 

We obtain \method neighborhood embeddings using three different configurations of edge modalities (1) Spatial distance only (\textbf{\method DIST}); (2) Mobility only (\textbf{\method MOB}); (3) Both spatial distance and mobility (\textbf{\method DIST+MOB}).
%as described in Section~\ref{sec: intra}. 
We compare the embedding with the one derived using \textbf{Urban2Vec} method~\cite{wang2020urban2vec}, which rely solely on intra-neighborhood modalities, and \textbf{GAE}~\cite{DBLP:journals/corr/KipfW16}, which extract information from mobility graph using Graph Autoencoder. 

\section{Results and Discussion}
\subsection{Predicting Demographics and Economics}
In this task, we treat trained neighborhood embeddings as input features to predict ACS demographic and economic attributes for each census tract. We choose Median Age, Years of Education, and Percentage of White Population as demographic attributes, and Poverty Rate, Average Household Income and Employment Rate as economic attributes. We apply PCA to reduce the embedding dimensions to 50 before running the regression model. In this work, we try both linear regression and random forest regression. Census tracts are split into training set (85\%), and test set (15\%). We use $R^2$ as the major metrics and randomly reshuffle train/test split for 20 rounds to estimate variance of the performance.

As is shown in Figure~\ref{fig:plot}, two models trained with single edge modality outperform one another on different attributes: For example, for Median Age and Years of Education, \textbf{\method DIST} outperforms \textbf{\method MOB}, while \textbf{\method MOB} has a higher average $R^2$ for Percentage of White Population and Employment Rate.
However, by combing both modalities, \textbf{\method DIST + MOB} always outperform both of them and the baseline models \textbf{Urban2Vec} and \textbf{GAE} on all demographic and economic attributes, indicating the benefits of incorporating both intra- and inter-neighborhood modalities to capture mult-perspective urban characteristics.
Linear regression results from Table~\ref{table:demographics} follow a similar pattern: \textbf{\method DIST+MOB} outperforms all other models on all attributes except Percentage of White Population. 

\subsection{Training with Multi-City Data}
Since we adopt a contrastive sampling approach to learn the graph structure, we can easily scale up experiments to multiple cities without facing any memory issue. In this experiment, we investigate the improvements from training with merged data of both Chicago and New York City. Table~\ref{table:multi-city} shows the mean of $R^2$ for predicting all 6 demographic and economic attributes using linear regression. 
As is shown, using multi-city training set in Chicago yields better prediction performance but not for New York City. This may be  explained by the relative sparse mobility data in New York City. 
\begin{table}[h!]
    \centering
    \resizebox{0.4\textwidth}{!}{%
    \begin{tabular}{l|c|c|c}
      \toprule
      \multirow{2}{*}{Model}  & \multirow{2}{*}{Training set}  & \multicolumn{2}{|c}{Test set} \\
       \cmidrule{3-4} 
                              &                                & Chicago & New York City\\ 
      \midrule
    %   POI+SV & Combined - Single City & +0.027 & +0.005\\
    %   \multirow{2}{*}{POI+SV} & Single city & $0.479$ & $0.471$\\ 
    %                           & Combined & $0.506$ & $0.476$\\ 
    %   \midrule
    %   POI+SV+Mobility & Combined - Single City & +0.014 & -0.006\\
         \multirow{2}{*}{\method MOB} & Single-city & $0.613$ & $0.524$ \\
                                      & Multi-city & $0.627$ & $0.518$ \\
      \bottomrule
    \end{tabular}
    }
    \caption{Average prediction $R^2$, training on single-/multi-city data.}
    %\caption{Average linear regression prediction $R^2$ of multi-city training}
    \label{table:multi-city}
\end{table}

\subsection{Qualitative Analysis of the Embedding Space}
\subsubsection{Clustering of Neighborhood Embeddings}
% Kmeans census track embedding into 10 clusters
% Use the cluster as color hue
% Plot the census tract on the map 

\begin{figure}[ht!]
  \centering
  \includegraphics[width=0.43\textwidth]{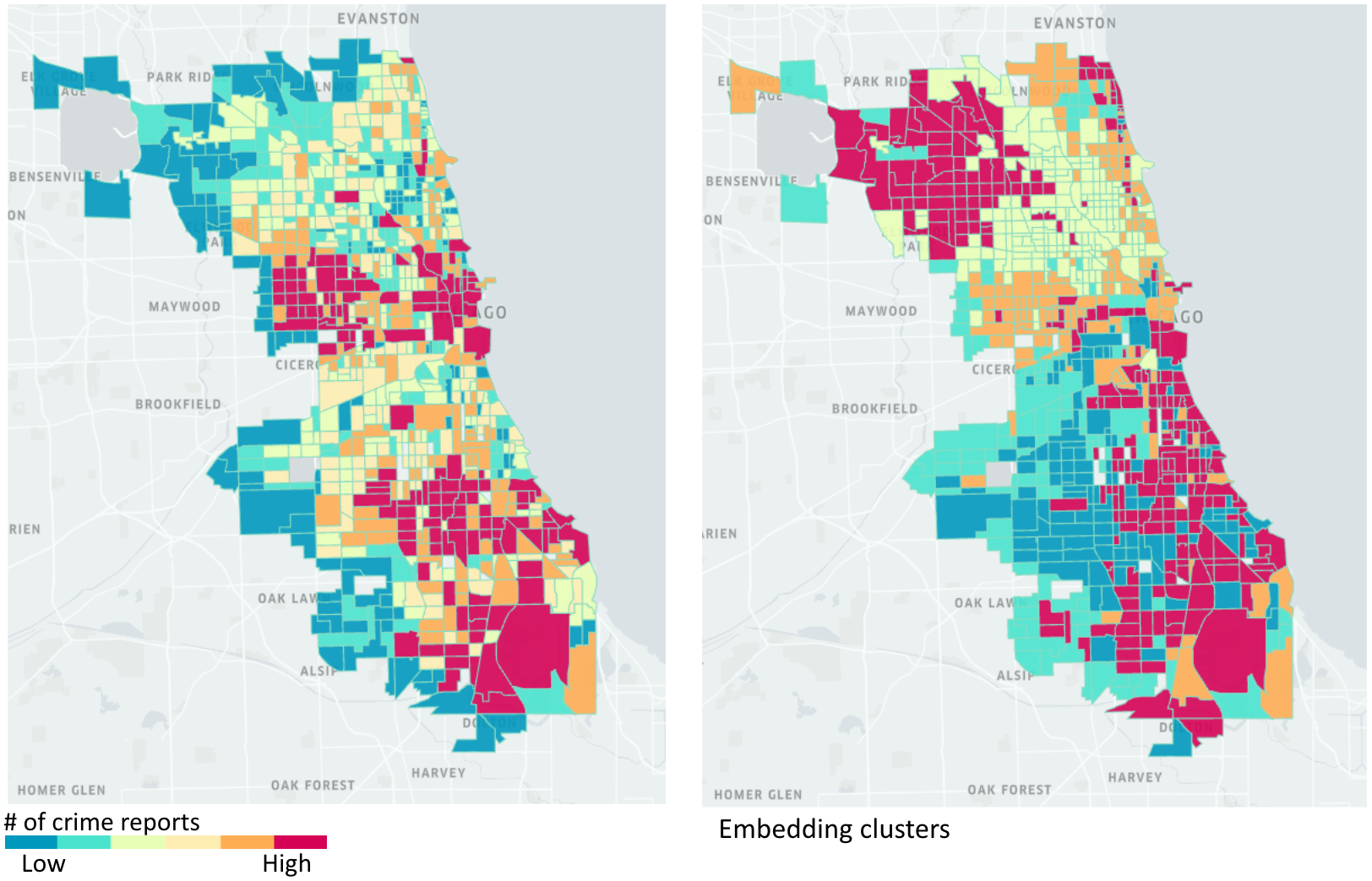}  
%   \td{Light background? Zoom in a bit?Add census tract geometry}
  \caption{Color-coded map based on {\bf Left:} Total number of crimes {\bf Right}: Embedding clusters derived by $k$-means ($k$=6) for Chicago.}
  \label{fig:kmeans}
  \centering
\end{figure}
To interpret the neighborhood embeddings learned from our models, we apply $k$-means clustering on the generated embedding. Figure~\ref{fig:kmeans} shows the results for $k=6$ in Chicago. As the plot shows, Downtown Chicago and South Chicago, which have a high number of crime reports\footnote{Chicago crime data available at \url{https://www.chicago.gov/city/en/dataset/crime.html}}, are clustered into one group (red), while neighborhoods in the north like Evanston are clustered into other groups (yellow and orange).

\subsubsection{Correlation with Geospatial and Mobility Proximity}
\begin{figure}[ht!]
  \centering
    
  \includegraphics[width=0.23\textwidth]{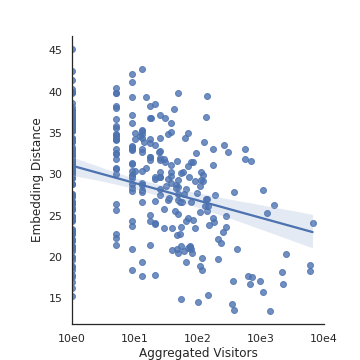} 
  \includegraphics[width=0.22\textwidth]{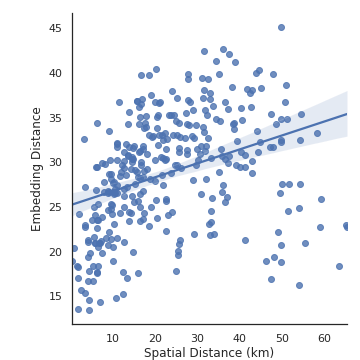} 
  \caption{Correlation between geospatial/mobility proximity of node pairs in the graph and the corresponding embedding distance in Chicago. {\bf Left}: The horizontal axis is the total number of visitors (bidirectional) between each pair from January 2018 to December 2019. {\bf Right}: The horizontal axis is the spatial distance measured in km. }
  \label{fig:correlation}
\end{figure}

In this analysis, we investigate the correlations between inter-neighborhood embedding distance and
their real-world proximity in terms of geo-distance or mobility. In Figure~\ref{fig:correlation}, we sample 0.1\% of the 1.6 M pairs of census tracts in Chicago and measure the L2 distances between their embedding vectors.  With a larger number of aggregated visitors in between, neighborhoods tend to have representations closer in the embedding space; as spatial distance becomes larger, two neighborhoods tend to fall further apart in the embedding space.
Such trends demonstrate that the embedding indeed captures both the geospatial and mobility relations through training.
% Calculate mobility distance/spatial distance between any pair of neighborhood
% Calculate embedding distance between any pair of neighborhood
% Plot the correlation 

\subsubsection{Neighborhood Embedding and Input Data Embedding}
\begin{figure}[ht!]
  \centering
  \includegraphics[width=0.45\textwidth]{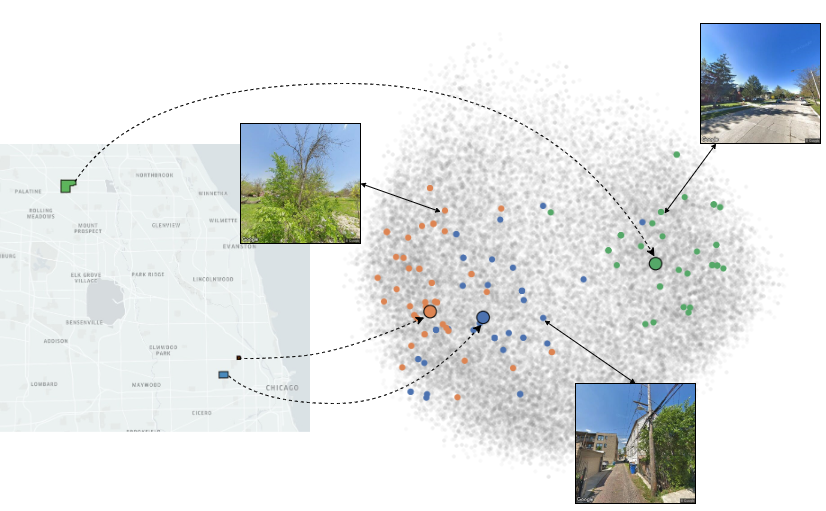} 
  \caption{Positions of embeddings in the plane of the first two PCA components, for both neighborhood and street view images. }
  \label{fig:container}
\end{figure}

We are also interested in whether the neighborhood embedding incorporates information from the geo-tagged point data. We apply PCA to extract the first two principal components of the embeddings of both neighborhoods and street views and plot their distribution in Figure~\ref{fig:container}. 
Large points with black borders denote neighborhoods; small points denote street view images, with the color indicating the neighborhood they belong to. Here, we randomly selected three census tracts for visualization.
Census tracts in {\bf \color{orange} Orange}, {\bf \color{BlueViolet} Blue}, and {\bf \color{ForestGreen} Green} have average household income of \$34,407, \$43,836, and \$113,479, respectively. 
As the plot shows, street view embeddings scatter around their corresponding neighborhood embedding. Though all three sampled images contain large portion of vegetation, their visual difference (e.g. trimmed or not, road landscape) can be reflected by their proximity in embedding space. 

% Get 2-3 neighborhood, and all the images, POIs in them 
% Dimension reduction and plot them

%17031240700
%17031231500
%17041240700

\section{Conclusion}
In this work, we develop \method, a framework to model urban neighborhoods as a multi-modal multi-graph and thus learn the neighborhood representation.
To demonstrate our framework, we use street view images and POIs as two modalities of data inside the neighborhood and both geospatial proximity and mobility pattern as two modalities of ``edges'' between neighborhoods.
We show the neighborhood embedding from our framework outperforms the ones from other multi-modal models in the downstream prediction tasks while preserving both proximity/mobility connections between neighborhoods, and relations between the neighborhood and street views.
The method we propose here is a general framework to learn representation for a graph with multi-modal ``node'' and multi-modal ``edge''. 
Such a framework can further integrate other modalities like satellite imagery (as components of the ``nodes'') and inter-region transactions (as ``edges''), and even be extended to learn the representation of other graph-structured data such as websites, which will be an important task in our future work.

\bibliographystyle{named}
\bibliography{ref}

\clearpage
\appendix

% \subsection{Additional Experiment Results}
\begin{table*}[h!]
    \centering
    \resizebox{0.95\textwidth}{!}{%
    \begin{tabular}{l|c|c|c|c|c|c}
      \toprule
        \multirow{2}{*}{Model} & \multicolumn{3}{|c}{Demographic characteristics} & \multicolumn{3}{|c}{Economic characteristics} \\
          \cmidrule{2-7} 
        &  Median age & Years of education & Percentage of white population & Poverty rate & Average household income & Employment rate\\ 
      \midrule
      Urban2Vec \cite{wang2020urban2vec}& $4.081$ & $0.724$  & $0.186$ & $0.076$ & $20,270$ & $0.047$\\
      GAE \cite{DBLP:journals/corr/KipfW16}& $4.283$ & $0.740$ & $0.182$ & $0.073$& $20,531$ & $0.046$\\
      \method DIST & $3.983$ & $0.642$ & $0.153$ & $0.072$ & $19,295$ & $0.043$\\
      \method MOB & $3.975$ & $0.600$ & $\mathbf{0.128}$& $0.064$ & $17,794$ & $0.039$\\
      \method DIST+MOB & $\mathbf{3.861}$ & $\mathbf{0.583}$ & $0.129$ & $\mathbf{0.064}$ & $\mathbf{17,509}$ & $\mathbf{0.038}$\\
      \bottomrule
    \end{tabular}
    }
    \caption{Prediction MAE on demographic and economic attributes with linear regression model in Chicago}
    \label{table:mae_linear}
\end{table*}
\begin{table*}[h!]
    \centering
    \resizebox{0.95\textwidth}{!}{%
    \begin{tabular}{l|c|c|c|c|c|c}
      \toprule
        \multirow{2}{*}{Model} & \multicolumn{3}{|c}{Demographic characteristics} & \multicolumn{3}{|c}{Economic characteristics} \\
          \cmidrule{2-7} 
        &  Median age & Years of education & Percentage of white population & Poverty rate & Average household income & Employment rate\\ 
      \midrule
      Urban2Vec \cite{wang2020urban2vec}& $4.181$ & $0.739$  & $0.193$ & $0.079$ & $18,728$ & $0.048$\\
      GAE \cite{DBLP:journals/corr/KipfW16}& $4.104$ & $0.716$ & $0.140$ & $0.070$& $18,693$ & $0.041$\\
      \method DIST & $3.747$ & $0.608$ & $0.140$ & $0.064$ & $16,493$ & $0.039$\\
      \method MOB & $4.014$ & $0.690$ & $0.114$& $0.064$ & $17,088$ & $0.036$\\
      \method DIST+MOB & $\mathbf{3.716}$ & $\mathbf{0.587}$ & $\mathbf{0.064}$ & $\mathbf{0.064}$ & $\mathbf{15,578}$ & $\mathbf{0.035}$\\
      \bottomrule
    \end{tabular}
    }
    \caption{Prediction MAE on demographic and economic attributes with random forest model in Chicago}
    \label{table:mae_rf}
\end{table*}
\begin{table*}[h!]
    \centering
    \resizebox{0.95\textwidth}{!}{%
    \begin{tabular}{l|c|c|c|c|c|c}
      \toprule
        \multirow{2}{*}{Model} & \multicolumn{3}{|c}{Demographic characteristics} & \multicolumn{3}{|c}{Economic characteristics} \\
          \cmidrule{2-7} 
        &  Median age & Years of education & Percentage of white population & Poverty rate & Average household income & Employment rate\\ 
      \midrule
      Urban2Vec \cite{wang2020urban2vec}& $0.430$ & $0.634$  & $0.496$ & $0.494$ & $0.533$ & $0.508$\\
      GAE \cite{DBLP:journals/corr/KipfW16}& $0.419$ & $0.648$ & $0.512$ & $0.510$& $0.557$ & $0.529$\\
      \method DIST & $0.453$ & $0.680$ & $0.572$ & $0.523$ & $0.580$ & $0.544$\\
      \method MOB & $0.450$ & $0.702$ & $0.614$& $0.568$ & $0.617$ & $0.579$\\
      \method DIST+MOB & $\mathbf{0.472}$ & $\mathbf{0.717}$ & $\mathbf{0.618}$ & $\mathbf{0.572}$ & $\mathbf{0.627}$ & $\mathbf{0.584}$\\
      \bottomrule
    \end{tabular}
    }
    \caption{Prediction Kendall's $\tau$ on demographic and economic attributes with linear regression model in Chicago}
    \label{table:tau_linear}
\end{table*}
\begin{table*}[h!]
    \centering
    \resizebox{0.95\textwidth}{!}{%
    \begin{tabular}{l|c|c|c|c|c|c}
      \toprule
        \multirow{2}{*}{Model} & \multicolumn{3}{|c}{Demographic characteristics} & \multicolumn{3}{|c}{Economic characteristics} \\
          \cmidrule{2-7} 
        &  Median age & Years of education & Percentage of white population & Poverty rate & Average household income & Employment rate\\ 
      \midrule
      Urban2Vec \cite{wang2020urban2vec}& $0.398$ & $0.618$  & $0.455$ & $0.473$ & $0.546$ & $0.485$\\
      GAE \cite{DBLP:journals/corr/KipfW16}& $0.414$ & $0.632$ & $0.581$ & $0.502$& $0.579$ & $0.548$\\
      \method DIST & $0.487$ & $0.694$ & $0.603$ & $0.569$ & $0.619$ & $0.582$\\
      \method MOB & $0.436$ & $0.658$ & $0.642$& $0.556$ & $0.631$ & $0.596$\\
      \method DIST+MOB & $\mathbf{0.493}$ & $\mathbf{0.711}$ & $\mathbf{0.673}$ & $\mathbf{0.567}$ & $\mathbf{0.648}$ & $\mathbf{0.624}$\\
      \bottomrule
    \end{tabular}
    }
    \caption{Prediction Kendall's $\tau$ on demographic and economic attributes with random forest model in Chicago}
    \label{table:tau_rf}
\end{table*}
% \subsection{Dataset Statistics}
\begin{table*}[h!]
    \centering
    \resizebox{0.5\textwidth}{!}{%
    \begin{tabular}{l|c|c|c}
      \toprule
       & \# Street views & \# POIs & \# Neighborhoods (census tract)\\ 
      \midrule
      Chicago & $64,739$ & $38,445$ & $1,294$\\ 
      New York City & $67,271$ & $50,697$ & $1,310$\\
      \bottomrule
    \end{tabular}
    }
    \caption{Street views and POI data statistics}
    \label{table:stats}
\end{table*}
\end{document}